\documentclass[conference]{IEEEtran}
\IEEEoverridecommandlockouts
\usepackage{cite}
\usepackage{amsmath,amssymb,amsfonts}
\usepackage{algorithmic}
\usepackage{graphicx}
\usepackage{textcomp}
\usepackage{array}  
\usepackage[table]{xcolor}
\usepackage{tabularx}
\usepackage{booktabs}
\usepackage{listings}
\usepackage[compatibility=false]{caption}
\usepackage{subfiles}
\usepackage{tikz}
\usepackage{float}
\usepackage{multirow}
\usepackage{multicol}
\usetikzlibrary{positioning, arrows.meta}
\usepackage{verbatim}
\usepackage[english]{babel}
\usepackage[autostyle, english = american]{csquotes}
\MakeOuterQuote{"}
\usepackage{listings}
\usepackage{xcolor}
\usepackage{url}
\lstdefinelanguage{VCD}{
    keywords={timescale, scope, var, upscope, enddefinitions, dumpvars, end, module, wire},
    keywordstyle=\color{blue}\bfseries,
    sensitive=true,
    comment=[l]{\$comment},
    morecomment=[s]{\$comment}{\$end},
    commentstyle=\color{green!60!black},
    breaklines=true,
    showstringspaces=false
}
\colorlet{punct}{red!60!black}
\definecolor{background}{HTML}{EEEEEE}
\definecolor{delim}{RGB}{20,105,176}
\colorlet{numb}{magenta!60!black}
\lstdefinelanguage{json}{
    showstringspaces=false,
    breaklines=true,
    literate=
     *{0}{{{\color{numb}0}}}{1}
      {1}{{{\color{numb}1}}}{1}
      {2}{{{\color{numb}2}}}{1}
      {3}{{{\color{numb}3}}}{1}
      {4}{{{\color{numb}4}}}{1}
      {5}{{{\color{numb}5}}}{1}
      {6}{{{\color{numb}6}}}{1}
      {7}{{{\color{numb}7}}}{1}
      {8}{{{\color{numb}8}}}{1}
      {9}{{{\color{numb}9}}}{1}
      {:}{{{\color{punct}{:}}}}{1}
      {,}{{{\color{punct}{,}}}}{1}
      {\{}{{{\color{delim}{\{}}}}{1}
      {\}}{{{\color{delim}{\}}}}}{1}
      {[}{{{\color{delim}{[}}}}{1}
      {]}{{{\color{delim}{]}}}}{1},
}
\newcolumntype{Y}{>{\centering\arraybackslash}X}
\def\BibTeX{{\rm B\kern-.05em{\sc i\kern-.025em b}\kern-.08em
    T\kern-.1667em\lower.7ex\hbox{E}\kern-.125emX}}

\newcommand{\blfootnote}[1]{%
  \begingroup
  \renewcommand\thefootnote{}\footnote{#1}%
  \addtocounter{footnote}{-1}%
  \endgroup
}

\begin{document}
\title{WaveformQA: Benchmarking LLM Temporal Reasoning on Digital Waveforms}

\author{\IEEEauthorblockN{Yichuan Liu\textsuperscript{*}}
\IEEEauthorblockA{\textit{Tenstorrent} \\
\textit{Austin, TX, USA}\\
yichuanliu@tenstorrent.com \\
}
\and
\IEEEauthorblockN{Daniel Cummings\textsuperscript{*}}
\IEEEauthorblockA{\textit{Tenstorrent} \\
\textit{Austin, TX, USA}\\
dcummings@tenstorrent.com \\
}
\and
\IEEEauthorblockN{Nick Vadlamudi}
\IEEEauthorblockA{\textit{Tenstorrent} \\
\textit{Austin, TX, USA}\\
nvadlamudi@tenstorrent.com \\
}}

\maketitle
\blfootnote{\copyright~2026 IEEE. Personal use of this material is
permitted. Permission from IEEE must be obtained for all other uses, in
any current or future media, including reprinting/republishing this
material for advertising or promotional purposes, creating new collective
works, for resale or redistribution to servers or lists, or reuse of any
copyrighted component of this work in other works.}

\renewcommand\thefootnote{*}\footnotetext{These authors contributed equally.}

\begin{abstract}
Large Language Models (LLMs) have demonstrated strong capabilities in code generation and reasoning, yet their ability to perform temporal reasoning over digital waveform data remains largely unexplored. Although reasoning over digital waveforms is a critical bottleneck in design verification, existing benchmarks primarily evaluate hardware description language (HDL) code generation and use waveforms only as supplementary context. This paper presents WaveformQA, an open-source question-answering benchmark for evaluating LLM temporal reasoning over digital waveforms. The benchmark comprises 360 questions with programmatically generated ground truths across eight categories of varying difficulty, including questions targeting multi-signal correlation and event ordering. Waveforms are generated from open-source design implementations, ensuring reproducibility and grounding the benchmark in real hardware behavior. Evaluation of frontier LLMs reveals that while models achieve reasonable accuracy on simple queries, performance degrades due to context window limitations and reasoning difficulties on complex temporal and multi-step questions. In addition, we show that an event-time JSON representation of waveforms improves LLM reasoning accuracy versus the standardized value change dump (VCD) format. The open-source framework supports extending to new question categories and importing new waveform sources, enabling researchers to rapidly prototype temporal reasoning experiments. 
\end{abstract}

\begin{IEEEkeywords}
LLM-aided Design, Benchmark, Temporal Reasoning, Design Verification, Electronic Design Automation
\end{IEEEkeywords}

\section{Introduction}

The rapid advancement of Large Language Models (LLMs) has transformed software engineering, mathematical reasoning, and code generation tasks. As these capabilities mature, there is growing interest in applying LLMs to electronic design automation (EDA), where AI assistants can be leveraged to accelerate hardware verification, debugging, and design space exploration \cite{chipbench,assertllm}. A major bottleneck in EDA is design verification, which consumes approximately 60\% of total project time and is a primary cause of costly, delayed silicon re-spins \cite{foster2019wilson8}. A central pain point in this process is simulation waveform analysis: verification engineers must carefully inspect the timing relationships among multiple, intertwined signals across large execution traces. Engineers typically use waveform viewers (e.g., GTKWave \cite{Fuest2026gtkwave}) to visually inspect simulation traces from formats such as the value change dump (VCD). In complex designs, these intricate multi-signal interactions are still largely analyzed manually, making the process especially time-consuming~\cite{fvdebug}. 

However, it remains largely unexplored whether LLMs possess the precise temporal reasoning capabilities needed to assist with waveform analysis. Although LLMs have proven adept at semantic and generative tasks like code generation, waveform analysis requires distinct skills such as correlating the behavior of hundreds of intertwined signals, computing exact timing relationships within precise time windows, and counting discrete events across four-state logic.

To date, there are no benchmarks to systematically evaluate the ability of LLMs to reason directly on digital waveform data. Current hardware-focused LLM benchmarks evaluate hardware description language (HDL) tasks and register transfer level (RTL) code generation, but not runtime trace analysis. Temporal reasoning benchmarks from natural language processing (NLP) and formal methods operate on fundamentally different data structures such as natural language text or event-ordering problems that lack the nanosecond-precision timing and multi-bit signals of hardware traces. This gap limits the measurement of whether LLMs can assist with the highly time-consuming aspect of verification workflows.

\begin{figure*}[!t]
    \centering
    \includegraphics[width=0.7\textwidth]{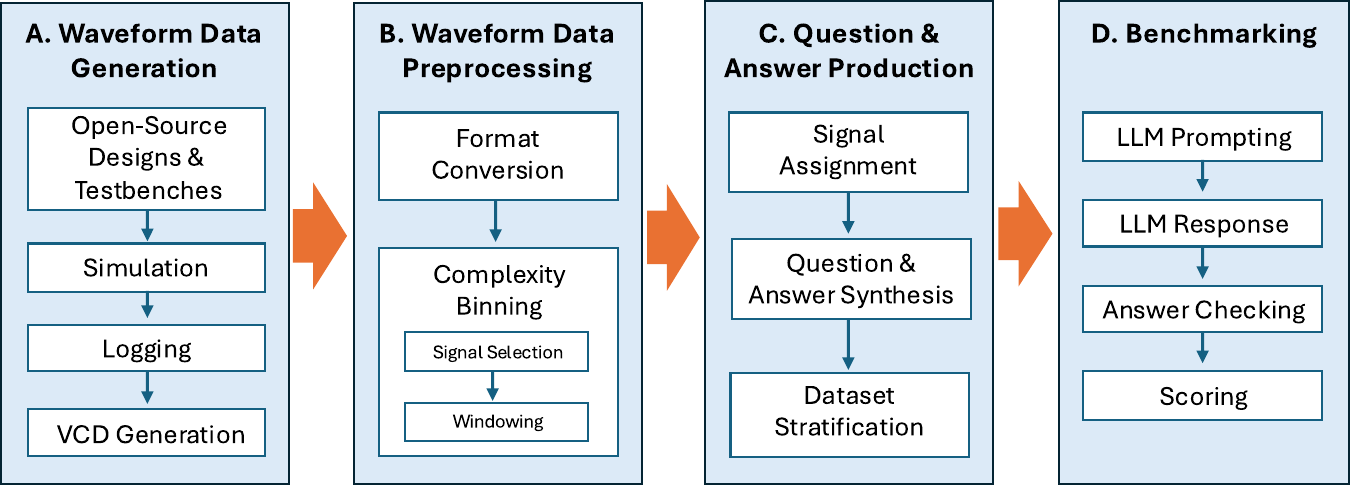}
    \caption{WaveformQA benchmarking workflow with four stages: (A) waveform data generation, (B) waveform data preprocessing, (C) question and answer production, and (D) benchmarking.}
    \label{fig:flowchart}
\end{figure*}

We present \textit{WaveformQA}, a question–answering benchmark specifically designed to evaluate the temporal reasoning capabilities of LLMs over digital waveforms from hardware simulations. Our main contributions are:

\begin{itemize}

\item A collection of real-world hardware traces derived from open-source RISC-V processor implementations (PicoRV32 \cite{picorv32}, DarkRISCV \cite{darkriscv}, Ibex \cite{ibex}, SERV~\cite{serv}, biRISC-V~\cite{biriscv}), capturing realistic simulation behavior with complex multi-signal interactions.

\item A suite of 360 targeted verification questions spanning 8 reasoning categories (e.g., temporal ordering, multi-step reasoning, correlation), designed to reflect tasks encountered by design verification engineers. Each question is associated with programmatically verified ground truth.

\item A waveform formatting study comparing VCD and JSON encodings, showing that the data format significantly affects LLM reasoning performance on waveforms.

\item An open-source evaluation framework supporting extensible benchmark generation, integration of new waveform sources, and evaluation across diverse LLM platforms.
\end{itemize}

Our evaluation of commercial and open source frontier models (Claude 4.5 \& 4.6 Sonnet \cite{claude45sonnet}, \cite{sonnet4_6}, Gemini 2.5 \cite{gemini25}, Qwen3 30B \cite{qwen3}) provides insight into which question types pose challenges for these models and demonstrates the importance of waveform data representation. WaveformQA establishes a principled framework for measuring progress in LLM-based waveform analysis as AI integration deepens in EDA.

\section{Related Work}

\subsection{Temporal Reasoning Benchmarks}

Temporal reasoning has been studied in NLP and formal verification, but existing benchmarks operate on fundamentally different data structures than digital waveform data. NLP benchmarks such as TimeQA \cite{timeqa}, TempQuestions \cite{tempquestions}, and TIMEBENCH \cite{timebench} evaluate qualitative temporal relations over vague natural language text. Formal verification benchmarks like LTLBench \cite{ltlbench} and TemporalBench \cite{tempobench} reason over abstract event-based sequences rather than specific signals with bit widths. TRAM \cite{tram} uses synthetic scenarios that lack precise numerical constraints and four-state logic. Digital waveforms are fundamentally different: they are highly multi-dimensional, require nanosecond-precision numerical timing, involve complex encoding (binary, hexadecimal, four-state values), and embed domain-specific semantics (control flow, data dependencies).

\subsection{Hardware and EDA Benchmarks}

Recent benchmarks evaluate LLM capabilities across the hardware design flow, but focus on code-level tasks (see Table \ref{tab:benchmark_comparison}). Benchmarks like RTLLM \cite{rtllm}, VerilogEval \cite{verilogeval}, and SVA-Eval \cite{assertllm} test HDL/assertion generation, while Meic \cite{meic}, Chrysalis \cite{chry}, Vert \cite{Vert} and FVDebug \cite{fvdebug} focus on code repair and debugging. Although some use waveforms as an auxiliary context, none systematically evaluates temporal reasoning over signal traces.

ChipBench \cite{chipbench}, a recent and comprehensive hardware AI benchmark, includes VCD waveforms as a supplemental context for debugging tasks, but found they often hurt performance: providing waveforms improved results on only 8 of 21 models tested, while degrading performance on the remainder. The authors conclude that "while waveform data contain critical debugging information, most current LLMs lack the capability to interpret it effectively," identifying waveform analysis tools and waveform-aware training as essential future work. However, ChipBench focuses on code-level debugging rather than temporal reasoning, and does not systematically evaluate waveform formatting or interpretation capabilities across question types or reasoning categories.

This RTL-centric focus leaves a critical gap: in practice, verification engineers dedicate the largest portion of their time to debug activities \cite{foster2019wilson8}, yet current LLM benchmarks predominantly target code generation rather than waveform analysis tasks which are central to debugging workflows. WaveformQA addresses this gap as the first benchmark to systematically evaluate LLM temporal reasoning over digital waveforms, with precise timing requirements, multi-signal correlation, automated ground truth verification, and stratification across question types.

\begin{table}[!t]
\centering
\caption{Hardware Verification LLM Benchmarks by Primary Evaluation Focus}
\label{tab:benchmark_comparison}

\renewcommand{\arraystretch}{1.1}
\begin{tabular}{p{0.38\columnwidth}p{0.5\columnwidth}}
\toprule
\textbf{Benchmark} & \textbf{Primary Evaluation Focus} \\
\midrule
\scriptsize
RTLLM, VerilogEval, SVA-Eval & HDL/assertion generation from specs \\
Meic, Chrysalis, Vert & RTL/assertion repair and refinement \\
FVDebug & RTL debugging assistance \\
ChipBench & RTL debugging (VCDs as context) \\
\midrule
\textbf{WaveformQA} & \textbf{Temporal reasoning over waveforms} \\
\bottomrule
\end{tabular}
\end{table}

\begin{table*}[t]
\centering
\setlength{\tabcolsep}{5pt}
\renewcommand{\arraystretch}{1}
\scriptsize
\caption{Generated Question Categories \& Sub-Categories}
\label{tab:question-description}
\begin{tabularx}{\textwidth}{@{} l l X X @{}}
\toprule
\textbf{Category} & \textbf{Sub-Category} & \textbf{Description} & \textbf{Example} \\
\midrule
\multirow{3}{*}{Lookup} & Value at Time & Signal value at specific time or edge & "Value of data\_out at 5th rising edge of clk?" \\
 & Value When Changes & Signal value when another transitions & "Value of addr when enable goes 0→1 for 4th time?" \\
 & First Match & Value at first occurrence of condition & "Value of data\_out first time ready=1 AND valid=1?" \\
\midrule
\multirow{3}{*}{Counting} & Transition Count & Count signal changes in time window & "How many times does state change between t=1000-5000?" \\
 & Rising Edge Count & Count edges, optionally gated & "Rising edges of clk while reset low, t=500-3000?" \\
 & Conditional Count & Count edges while signal in state & "Rising edges of interrupt while enable high?" \\
\midrule
\multirow{3}{*}{Arithmetic} & Period & Average period over interval & "Average period of clk\_div between 3rd-12th enable edges?" \\
 & Frequency & Frequency with unit conversion & "Frequency of clk in MHz between t=1000-8000?" \\
 & Time Difference & Time between edges of two signals & "Time between 3rd rising req and 2nd falling ack?" \\
\midrule
\multirow{3}{*}{Temporal} & Before/After & Order of signal changes vs reference & "Does data\_valid change before/after 4th strobe edge?" \\
 & Ordering & Sort signals by next change time & "Order by next change after t=2500: [reset, busy, done]" \\
 & Last Before & Last value before reference time & "Last value of instruction before 7th fetch\_valid edge?" \\
\midrule
\multirow{3}{*}{Multistep} & Cycles After Trigger & Value N cycles after trigger & "Value of pc 8 cycles after interrupt goes high?" \\
 & nth Event Value & Value at high-order event & "Value of mem\_addr at 18th falling edge of read\_en?" \\
 & Conditional Sequence & Chain conditions across time & "After 4th valid fall, data\_out at 1st strobe rise w/ enable=1?" \\
\midrule
\multirow{3}{*}{Correlation} & Dual Condition & Two signals match simultaneously & "Value of data\_bus 2nd time ready=1 AND valid=0?" \\
 & Triple Condition & Three signals match pattern & "Value of addr 5th time enable=1, ready=1, error=0?" \\
 & Which Changes First & Which signal changes last after trigger & "After start high, which changes last: [busy, done, error]?" \\
\midrule
\multirow{3}{*}{Misdirection} & nth Occurrence & High-order event with decoys & "Value of output at 9th rising edge of clock?" \\
 & After Specific Event & Event after ordinal event & "Value of data at 1st valid rise after 3rd ready fall?" \\
 & Ignore Glitch & Stable value ignoring transients & "Stable chip\_select between t=1000-3000, ignoring glitches?" \\
\midrule
\multirow{3}{*}{FSM} & State Sequence & Ordered states in time window & "Sequence of fsm\_state between t=500-2000?" \\
 & Next State & Next state after transition & "When fsm\_state leaves IDLE 5th time, next state?" \\
 & Return Count & Count state transitions in window & "Times fsm\_state goes BUSY→IDLE, t=1000-8000?" \\
\bottomrule
\end{tabularx}
\end{table*}

\section{Methodology}

Figure~\ref{fig:flowchart} illustrates the WaveformQA framework. The following subsections detail each stage and how the benchmark is constructed.

\subsection{Waveform Generation}

To ground the evaluation in real hardware behavior, we generate waveform data from five open source RISC-V cores spanning a range of architectural complexity (Table~\ref{tab:cores}). For each core, we compile custom firmware targeting diverse execution patterns: compute intensive (fibonacci, CoreMark), memory intensive (bubblesort), and branch intensive (Dhrystone). Simulation with Icarus Verilog \cite{iverilog} or Verilator \cite{verilator} produces 13 VCD traces ranging from 338 to 4,185 signals.

\begin{table}[!t]
\centering
\caption{RISC-V Cores Used for Trace Generation}
\label{tab:cores}
\renewcommand{\arraystretch}{1.15}
\begin{tabular}{llrr}
\toprule
\textbf{Core} & \textbf{ISA} & \textbf{Signals} & \textbf{Traces} \\
\midrule
PicoRV32~\cite{picorv32} & RV32I/E & 338 & 2 \\
DarkRISCV~\cite{darkriscv} & RV32I & 394 & 3 \\
SERV~\cite{serv} & RV32I & 957 & 2 \\
biRISC-V~\cite{biriscv} & RV32IM & 1,642--2,295 & 4 \\
Ibex~\cite{ibex} & RV32IMC & 4,185 & 2 \\
\midrule
\multicolumn{2}{l}{\textbf{Total}} & \textbf{338--4,185} & \textbf{13} \\
\bottomrule
\end{tabular}
\end{table}

\begin{figure*}[!t]
    \centering
    \includegraphics[width=\textwidth]{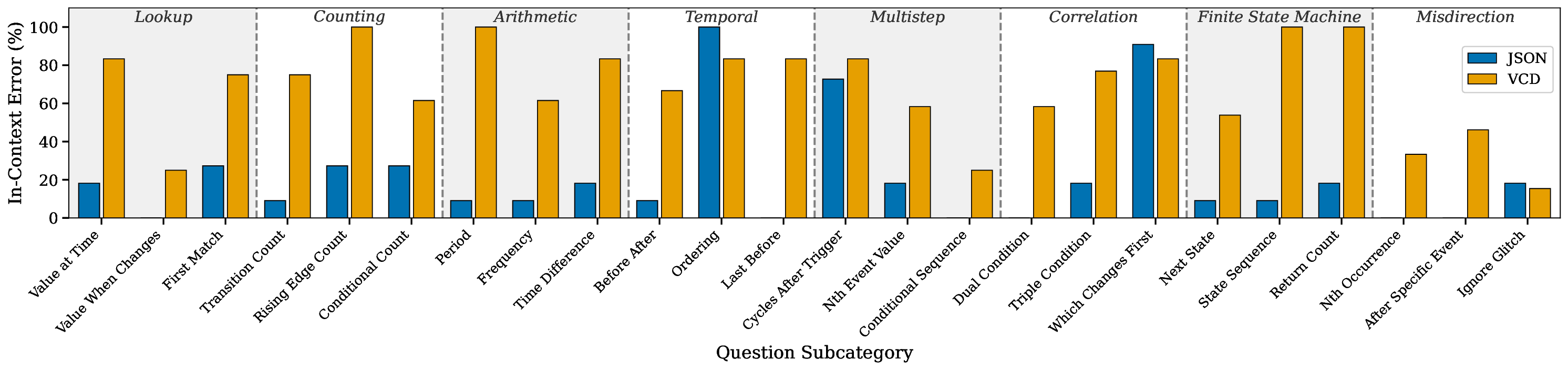}
    \caption{Paired bar chart of in-context error rate (\%) for JSON (teal) and VCD (gold) formatting, shown for each of the 24 question subcategories grouped by category. Evaluated on Gemini 2.5 Pro across all signal bins.}
    \label{fig:jsonvcd_accuracy_bar}
\end{figure*}

\subsection{Waveform Pre-Processing} 

Raw VCD traces undergo two major preprocessing stages: format conversion and complexity binning.

\textbf{Format conversion.} In the first stage, each trace is either kept in its original VCD format (IEEE 1364 standard \cite{IEEE1364-2005}) or converted to an event-time structured JSON representation. Both are \textit{event-based} formats: rather than storing signal values at every simulation timestep, they record only \textit{transitions}, which are individual value changes on a signal (e.g., a clock edge or a bus update). In this work, a transition means any time a signal waveform changes state (e.g., 0, 1, Z, X). More information and examples can be found in Appendix \ref{appendix:vcd_vs_json}.

\textbf{Complexity binning.}
In the second stage, we apply \textit{complexity binning} to control the waveform dataset size along two dimensions: signal count and transition count. These dimensions affect reasoning difficulty in distinct ways: more signals widen the search space the model must navigate, while more transitions lengthen the sequences it must parse. They also govern token consumption: in the event-based representations described above, each transition adds tokens to the prompt, and each signal contributes declaration overhead. Because our designs span 338 to 4{,}185 signals (Table~\ref{tab:cores}), signal count alone can produce a large difference in the prompt size. We control both dimensions via a signal selection step followed by a windowing step.

In the signal selection step, we group the 13 traces into three \textit{signal count bins} based on their design complexity: 0 to 1{,}000 signals, 1{,}000 to 3{,}000 signals, and 3{,}000 to 5{,}000 signals. In the windowing step, waveform windows are extracted for each signal bin at five \textit{max transition} thresholds: 5{,}000, 7{,}000, 15{,}000, 30{,}000, and 60{,}000. The max transition count caps the total number of recorded value changes in the window; all signals in the design are preserved, but the temporal extent is shortened so that the cumulative transition count does not exceed the threshold. This controls input size without removing signals. Combined with the three signal bins, this produces a $3 \times 5 = 15$ bin complexity grid. Each bin yields a set of selected signals for a given temporal window; we refer to these as \textit{processed waveform dataset}, and to the collection within a bin as the \textit{waveform dataset group}.

\subsection{Question and Answer Production}

For each waveform dataset group, questions and answers are generated in three stages: signal assignment, question and answer synthesis, and stratification.

\textbf{Signal assignment.} In the first stage, signals in each processed waveform dataset are assigned a type based on their waveform behavior. For example, one-bit signals with a stable period are assigned the \textit{clock} type. These assignments produce a parsed waveform data representation for question generation.

\textbf{Question and answer synthesis.} In the second stage, a question generator selects appropriate signals from the parsed waveform dataset representation, samples random time points, computes ground truth directly from the trace, and instantiates a natural-language question via Python f-strings. To cover a wide range of reasoning skills and difficulties, we implement eight question categories, each with three subcategories, yielding 24 subcategories in total (Table~\ref{tab:question-description}). A deterministic seed per (waveform dataset, question/answer) pair ensures full reproducibility. More examples can be found in Appendix \ref{appendix:q_prompts}.

\textbf{Stratification.} In the third stage, to ensure that no single processed waveform dataset or question subcategory dominates the evaluation, we generate 24 questions (one per subcategory) for each of the 15 complexity bins and distribute them evenly over the processed dataset groups. This yields $15 \times 24 = \mathbf{360}$ \textbf{total questions}, each paired with the respective ground truth and processed dataset.

\subsection{Benchmarking}

For each of the 360 questions, the LLM receives the respective processed waveform dataset alongside a single question and a system prompt. The responses are extracted and compared against the ground truth. Performance is measured using three metrics, where $C$ denotes correct, $W$ wrong, and $X$ context exceeded where $N$ is the total number of questions ($N = C + W + X$):

\begin{itemize}
\item \textbf{Aggregate Accuracy} (A-Acc) $= \frac{C}{C + W + X}$. Treats context exceeded as incorrect, reflecting end-to-end usability.
\item \textbf{In-Context Accuracy} (IC-Acc) $= \frac{C}{C + W}$. Considers only answerable questions, isolating reasoning ability from context limits.
\item \textbf{In-Context Error} (IC-Err) $= 1 - \text{IC-Acc}$. It is the complement of IC-Acc.
\end{itemize}

\begin{figure}[!tb]
    \centering
    \includegraphics[width=8.8cm]{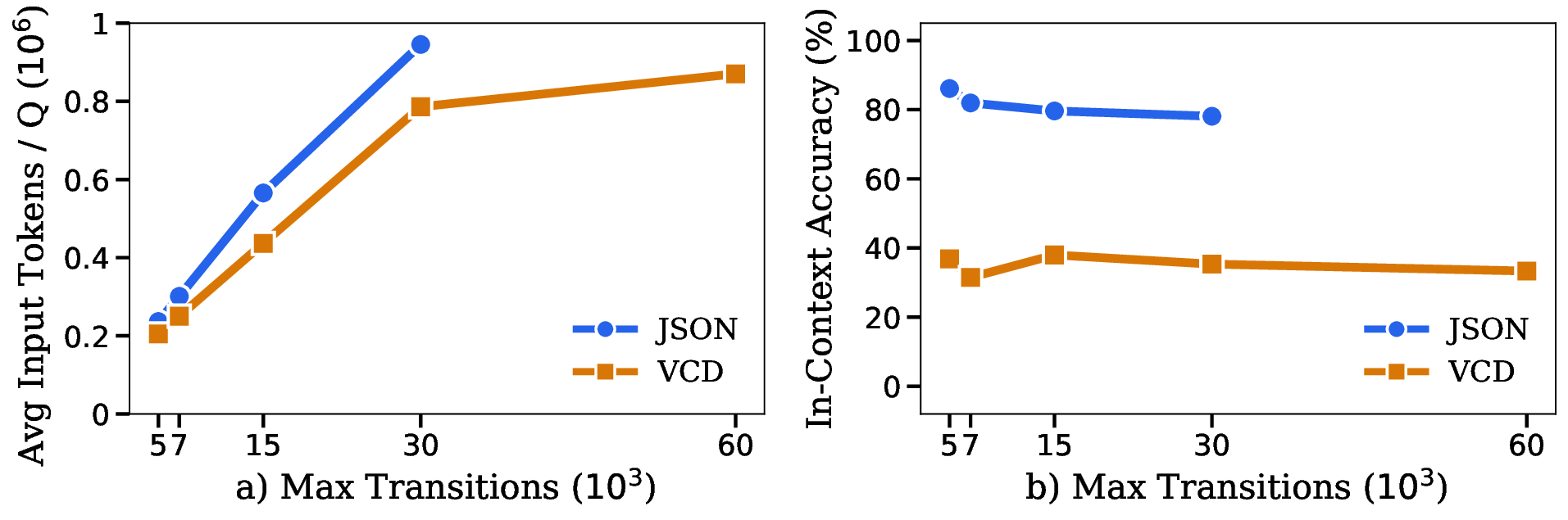}
    \caption{JSON vs.\ VCD on Gemini 2.5 Pro, averaged over all signal bins. (a)~Average input tokens per question ($\times 10^{6}$) vs.\ max transitions for the same two formats. (b)~In-context accuracy (\%) vs.\ max transitions ($\times 10^{3}$) for JSON and VCD.}
    \label{fig:jsonvcd_accuracy}
\end{figure}

\section{Experimental Results}

We first compare VCD and JSON waveform formats, then evaluate four frontier models across the complexity grid, and finally analyze accuracy by question category and subcategory as well as error type.

\subsection{Impact of Waveform Format}

Figure~\ref{fig:jsonvcd_accuracy}(a) shows average input tokens per question at each max-transition threshold (averaged across all signal bins), and Figure~\ref{fig:jsonvcd_accuracy}(b) shows in-context accuracy at the same thresholds. Both panels compare JSON and VCD representation formats evaluated on Gemini 2.5 Pro.

In terms of token efficiency, VCD consistently requires 15--30\% fewer tokens per question than JSON at transition thresholds where both formats fit within Gemini's context window. This compactness advantage lets VCD encode larger traces: at 60,000 transitions, VCD still fits within the 1M token limit while JSON does not.

Despite this token disadvantage, the JSON event-time format substantially outperforms VCD in accuracy and exhibits fewer systematic errors. From Figure~\ref{fig:jsonvcd_accuracy}(b), we observe an accuracy gain of 37--53\% over raw VCD at transition thresholds where both formats fit within the context window. This is further shown in Figure~\ref{fig:jsonvcd_accuracy_bar}, which presents the per-subtype accuracy for both formats across all 24 subcategories. VCD exhibits an average in-context error rate of 67.9\%, more than three times that of JSON (21.2\%), with VCD errors higher in 20 of 24 subcategories by more than 20 \%. This performance gap could occur because frontier models encountered more JSON formatted training during pre-training given its prevalence in web and API corpora~\cite{Tegridy_2025}. Alternatively, the JSON formatting may have also reduced the lookup overhead by preserving more explicit naming conventions compared to VCD. For example, the "clk" signal in JSON would be denoted as "!" in VCD, which invokes the LLM to have to increase overhead to better understand the signal. Given that we prioritize accuracy in this benchmark, all subsequent results use event-time JSON formatting.

\textbf{Key takeaway 1:} \textit{Structured event-time JSON waveform representations yield an accuracy gain of 37--53\% over raw VCD, at a 15--30\% token overhead.}

\begin{figure}[!t]
    \centering
    \includegraphics[width=8.8cm]{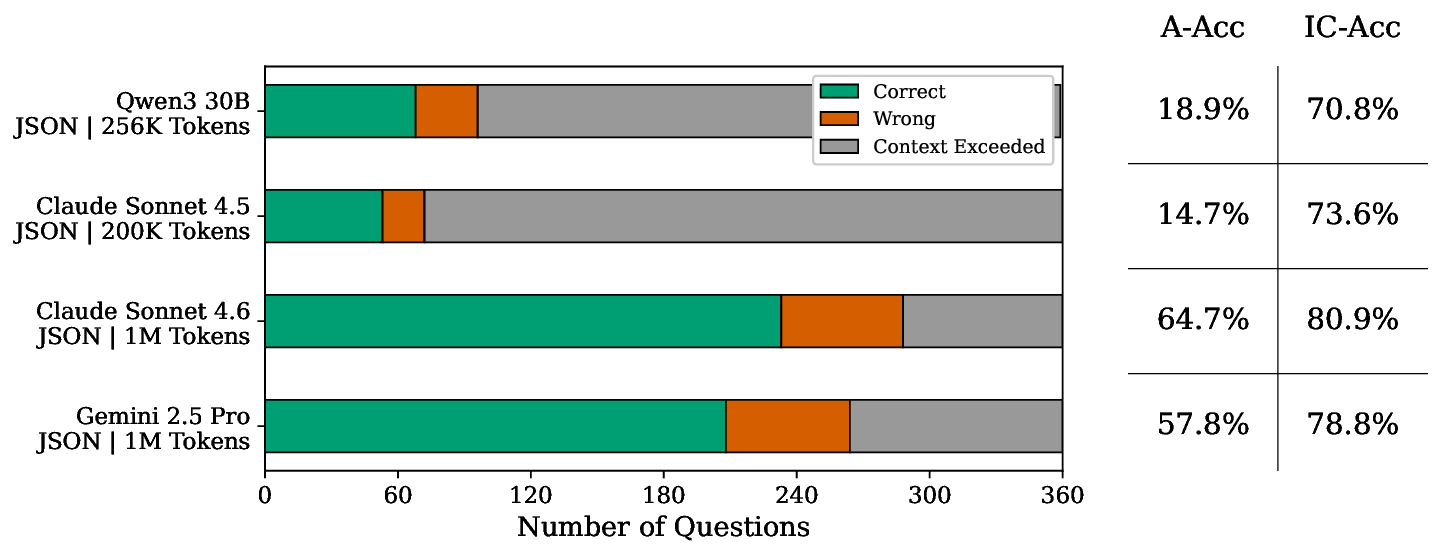}
    \caption{Bar chart showing the number of correct (green), wrong (orange), and context-exceeded (gray) responses out of 360 questions for each of the four models (JSON format). Aggregate accuracy (A-Acc) and in-context accuracy (IC-Acc) are listed to the right of each bar.}
    \label{fig:total_by_model}
\end{figure}

\begin{figure*}[!t]
    \centering
    \includegraphics[width=\textwidth]{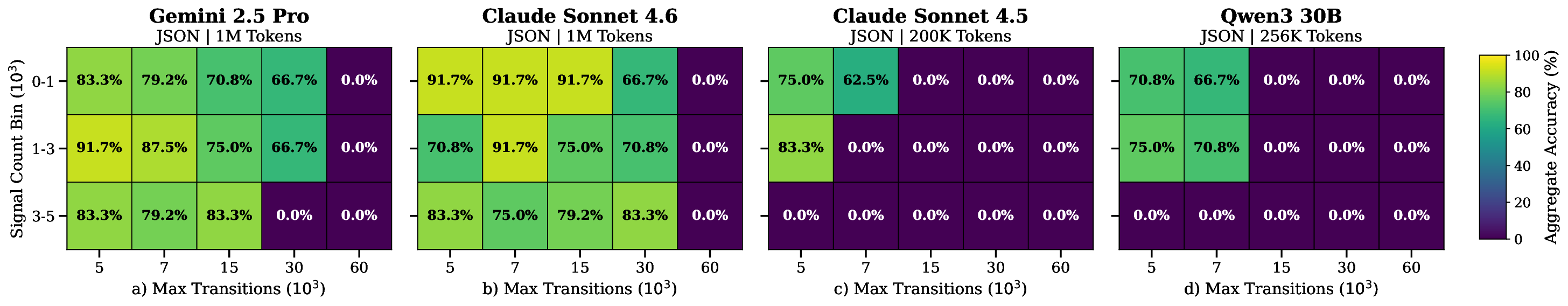}
    \caption{Heatmaps of aggregate accuracy (\%) across the $3 \times 5$ complexity grid for each model: (a)~Gemini 2.5 Pro (1M tokens), (b)~Claude Sonnet 4.6 (1M tokens), (c)~Claude Sonnet 4.5 (200k tokens), (d)~Qwen3 30B (256k tokens). Rows correspond to signal-count bins ($\times 10^{3}$), columns to max-transition thresholds ($\times 10^{3}$). 0\% is due to the model's context limit errors. Color scale ranges from 0\% (purple) to 100\% (yellow).}
    \label{fig:heatmap_bymodel}
\end{figure*}

\begin{figure*}[!t]
    \centering
        \includegraphics[width=\textwidth]{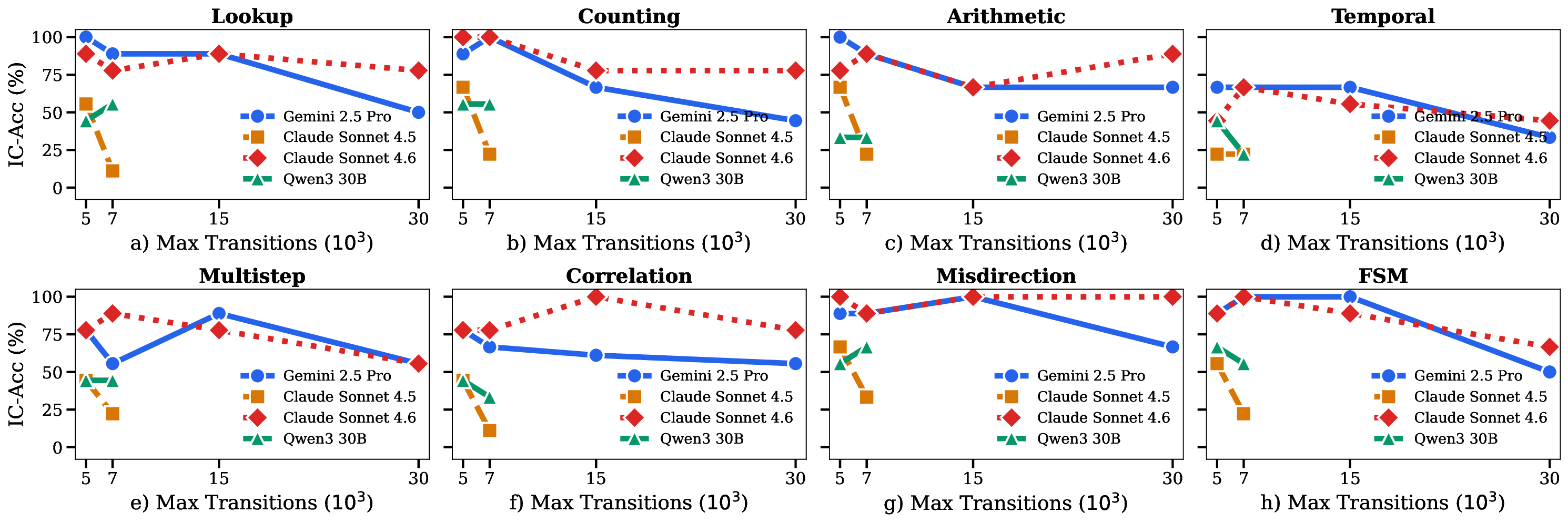}
    \caption{Average in-context accuracy (\%) vs.\ max transitions ($\times 10^{3}$) for each of the eight question categories: (a)~Lookup, (b)~Counting, (c)~Arithmetic, (d)~Temporal, (e)~Multistep, (f)~Correlation, (g)~Misdirection, (h)~Finite State Machine. Each subplot contains four lines, one per model: Gemini 2.5 Pro, Claude Sonnet 4.5, Claude Sonnet 4.6, and Qwen3 30B.}
    \label{fig:condensed_question_cat_vs_transitions}
\end{figure*}

\subsection{Performance by Model and Waveform Complexity}

We evaluate four models on all 360 questions using JSON format: Claude Sonnet 4.6 (1M token context), Gemini 2.5 Pro (1M token context), Qwen3 30B (256K token context), and Claude Sonnet 4.5 (200K token context).

Figure~\ref{fig:total_by_model} shows correct, wrong, and context exceeded responses for each model alongside A-Acc and IC-Acc. Claude Sonnet 4.6 (64.7\% A-Acc) and Gemini 2.5 Pro (57.8\%) answer the vast majority of questions, with only 20\% and 26.7\% exceeding their context limits respectively. In contrast, Qwen3 30B and Claude Sonnet 4.5 lose 73.3\% and 80\% of questions to context overflow, reducing A-Acc to 18.9\% and 14.7\%, respectively. This shows that aggregate accuracy is primarily driven by context window size.

The context bottleneck is also visible in Figure~\ref{fig:heatmap_bymodel}, which provides a finer breakdown of aggregate accuracy across the 15 complexity bins, each of which consist of a unique waveform dataset group. Claude Sonnet 4.5 fills only 3 of 15 bins before hitting context limits, Qwen3 30B fills 4 bins, Gemini fills 11 bins, and Claude Sonnet 4.6 fills 12 bins. However, context window size is not the sole factor. On the three bins that all four models can answer, Claude Sonnet 4.6 and Gemini both achieve 84.7\% IC-Acc, while Claude Sonnet 4.5 and Qwen3 30B reach only 73.6\% and 70.8\%. This 11 to 14 percentage point gap on identical questions reveals that the larger models also perform better independent of their context advantage.

Within answerable bins, two further patterns stand out. First, IC-Acc declines with transition count: Gemini drops from 86.1\% at 5k transitions to 78.1\% at 30k (averaged across signal bins), and Claude Sonnet 4.6 drops from 81.9\% to 73.6\% over the same range. This indicates that longer temporal sequences degrade reasoning performance even when the waveform dataset fits within the context window. Second, signal count has little effect on IC-Acc: Gemini averages 79.0\%, 84.4\%, and 81.9\% across all dataset bins, and Claude Sonnet 4.6 averages 85.4\%, 77.1\%, and 80.2\%. No consistent decline with signal count is observed, suggesting that widening the search space does not impair reasoning as long as the transition count remains comparable; signal count primarily increases token consumption, so its main impact is pushing traces toward the context limit sooner.

\textbf{Key takeaway 2:} \textit{In-context accuracy depends on transition count, declining 8 to 12 \% from 5k to 30k transitions, but not on signal count, which shows no consistent pattern across the three signal bins.}

\begin{figure*}[!t]
    \centering
    \includegraphics[width=\textwidth]{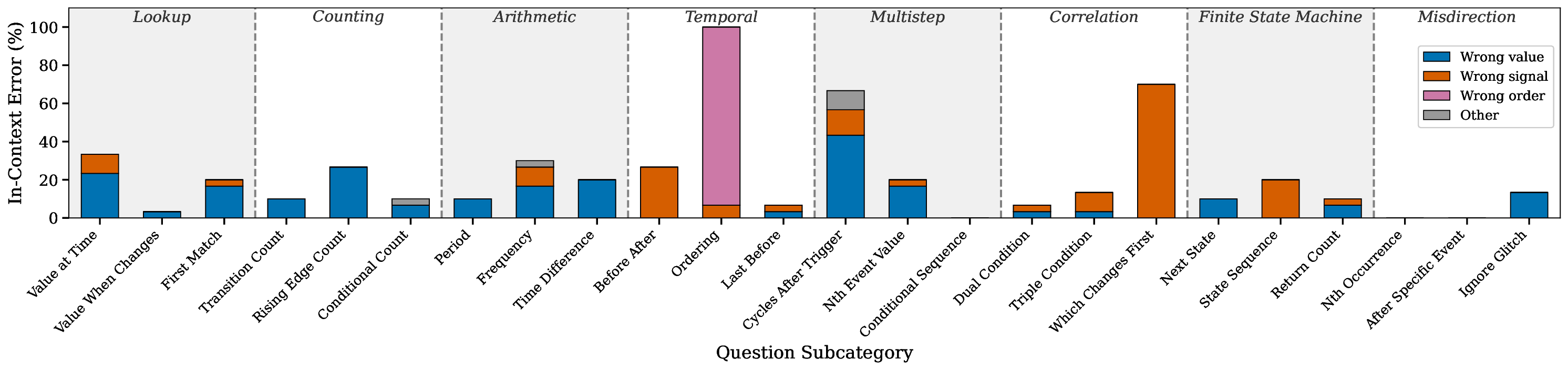}
    \caption{Bar chart of in-context error rate (\%) by question subcategory, aggregated across all four models. Bars are grouped by category and decomposed into four error types: wrong value (blue), wrong signal (orange), wrong order (pink), and other (gray).}
    \label{fig:error_analysis}
\end{figure*}

\subsection{Performance by Question Category and Subcategory}

Figure~\ref{fig:condensed_question_cat_vs_transitions}(a--h) shows in-context accuracy for each of the eight question categories, with one subplot per category and one line per model across transition thresholds. Overall, all models struggle most with \textit{Temporal}, \textit{Multistep}, and \textit{Correlation}, which consistently show lower accuracy across most transition thresholds compared to the remaining categories.

To understand what drives these category-level differences, Figure~\ref{fig:error_analysis} breaks down the in-context error by subcategory, aggregated across all four models, and decomposes each error by root cause: \textit{wrong value}, \textit{wrong signal}, \textit{wrong order}, or \textit{other}. Consistent with the category-level findings, the highest observed error rate subcategories are \textit{Ordering} (Temporal) with a near-total error rate of $\sim$95\%, followed by \textit{Which Changes First} (Correlation, $\sim$69\%) and \textit{Cycles After Trigger} (Multistep, $\sim$65\%). The error decomposition shows that \textit{wrong order} and \textit{wrong signal} errors dominate two of the most error-prone subcategories, indicating that event sequencing and signal identification are where models fail most severely. Numerical extraction errors (\textit{wrong value}) occur across all categories but at much lower rates. A per-model breakdown is provided in Appendix~\ref{appendix:error_model}.

\textbf{Key takeaway 3:} \textit{Temporal, Multistep, and Correlation categories are consistently the hardest across all models. The dominant errors in these categories are wrong order and wrong signal, indicating that event sequencing and multi-signal identification are fundamentally harder than numerical extraction for the evaluated LLMs.}

\section{Conclusion and Future Work}

We present WaveformQA, a novel benchmark for evaluating LLM temporal reasoning over digital waveforms. Evaluating four frontier models on 360 questions reveals three key findings: (1) Event-time JSON formatting outperforms VCD by 37 to 53 \% at a 15 to 30\% token overhead, (2) in-context accuracy declines 8 to 12 \% as transition count increases from 5k to 30k but appears unaffected by signal count, and (3) event sequencing and multi-signal identification were the hardest tasks across the models evaluated.

\textbf{Limitations.} Our traces cover RISC-V cores only, and not broader domains such as protocol controllers or memory subsystems. The 360-question scale identifies systematic failure modes but offers limited statistical power per subcategory.

\textbf{Future Work.} Key directions include (1) adding larger datasets for statistical significance and more challenging benchmark questions derived from this framework; (2) evaluating the visual modality using waveform viewer screenshots; (3) employing agentic context management via tool-calling APIs to avoid exhausting context windows; (4) scaling to more diverse hardware domains outside of RISC-V; and (5) exploring different waveform data representations such as time-grouped (rather than event-based) JSON formats.

\begin{flushleft}
\textbf{Code and data are available at:} \url{https://github.com/tenstorrent/waveformqa}
\end{flushleft}




\bibliographystyle{IEEEtran}
\bibliography{0_REFERENCES}

\section{Appendix}

\subsection{VCD versus JSON Representation Comparison} 
\label{appendix:vcd_vs_json}

To illustrate the representational trade-offs, consider a simple 4-bit counter, a fundamental building block in processor architectures (program counters, instruction fetch pipelines). This example circuit contains three signals: a 10ns-period clock (clk), an active-high reset (reset), and a 4-bit counter output (count[3:0]). The counter initializes to zero while reset is asserted. At t=5ns, reset deasserts and the counter begins incrementing on each rising clock edge: 0→1 at t=10ns, 1→2 at t=20ns, 2→3 at t=30ns.

\begin{lstlisting}[
  basicstyle=\ttfamily\scriptsize,
  columns=fullflexible,      % tightens inter-character spacing
  keepspaces=true,           % preserve VCD alignment
  breaklines=true,           % wrap any overlong line instead of overflowing
  breakatwhitespace=true,
  frame=single,
  framesep=3pt,
  xleftmargin=2pt,
  xrightmargin=2pt,
  captionpos=b,              % bottom caption sidesteps the top-rule overlap entirely
  caption={4-bit counter trace in VCD format},
  label={lst:vcd_example}
]
$timescale 1ns $end
$scope module counter $end
$var wire 1 ! clk $end
$var wire 1 " reset $end
$var wire 4 # count[3:0] $end
$upscope $end
$enddefinitions $end
$dumpvars
0!
1"
b0000 #
$end
#0
#5
1!
0"
#10
0!
b0001 #
#15
1!
#20
0!
b0010 #
#25
1!
#30
0!
b0011 #
\end{lstlisting}

VCD is the IEEE 1364 standard for recording signal transitions \cite{IEEE1364-2005}. VCD uses a compact delta-encoded representation where only state changes are logged. Signal declarations map readable names to single-character identifiers (\texttt{!}, ", \#), followed by timestamped value changes. The compactness of VCD is advantageous for storage efficiency, where large simulation traces compress well due to delta encoding. However, temporal queries require reconstructing signal state: answering \texttt{what is count at t=12ns?} requires parsing back to the most recent change before t=12ns (\texttt{b0001} at t=10ns). The cryptic single-character identifiers ! for clock, " for reset, \# for count) lack semantic context and become ambiguous in traces with hundreds of signals.

\begin{lstlisting}[
  basicstyle=\ttfamily\scriptsize,
  columns=fullflexible,      % tightens inter-character spacing
  keepspaces=true,           % preserve VCD alignment
  breaklines=true,           % wrap any overlong line instead of overflowing
  breakatwhitespace=true,
  frame=single,
  framesep=3pt,
  xleftmargin=2pt,
  xrightmargin=2pt,
  captionpos=b,              % bottom caption sidesteps the top-rule overlap entirely
  caption={4-bit counter trace in event-driven JSON format},
  label={lst:json_example}
]
{
  "trace_id": "counter_example",
  "time_range": [0, 30],
  "time_unit": "ns",
  "signals": {
    "clk": {
      "width": 1,
      "changes": [[0, 0], [5, 1], [10, 0], [15, 1],
                  [20, 0], [25, 1], [30, 0]]
    },
    "reset": {
      "width": 1,
      "changes": [[0, 1], [5, 0]]
    },
    "count": {
      "width": 4,
      "changes": [[0, "0x0"], [10, "0x1"],
                  [20, "0x2"], [30, "0x3"]]
    }
  }
}
\end{lstlisting}

Our structured JSON representation addresses VCD's interpretability limitations by grouping changes by signal name and including explicit metadata. The same 4-bit counter in JSON format as shown.

JSON provides explicit signal naming (``\texttt{clk}'' vs \texttt{!}), typed values with clear radix notation (\texttt{"0x1"} vs \texttt{b0001}), and structured access patterns (each signal's \texttt{changes} array is directly indexable). The \texttt{width} field disambiguates single-bit control signals from multi-bit buses. This verbosity costs 15--30\% more tokens per trace but yields 37--53\% accuracy gains in our evaluation (Section IV-A), as models leverage JSON's prevalence in pre-training corpora and avoid VCD parsing errors.

\subsection{Subcategory Error Breakdown per Model}
\label{appendix:error_model}

\begin{figure*}[!t]
    \centering
        \includegraphics[width=\textwidth]{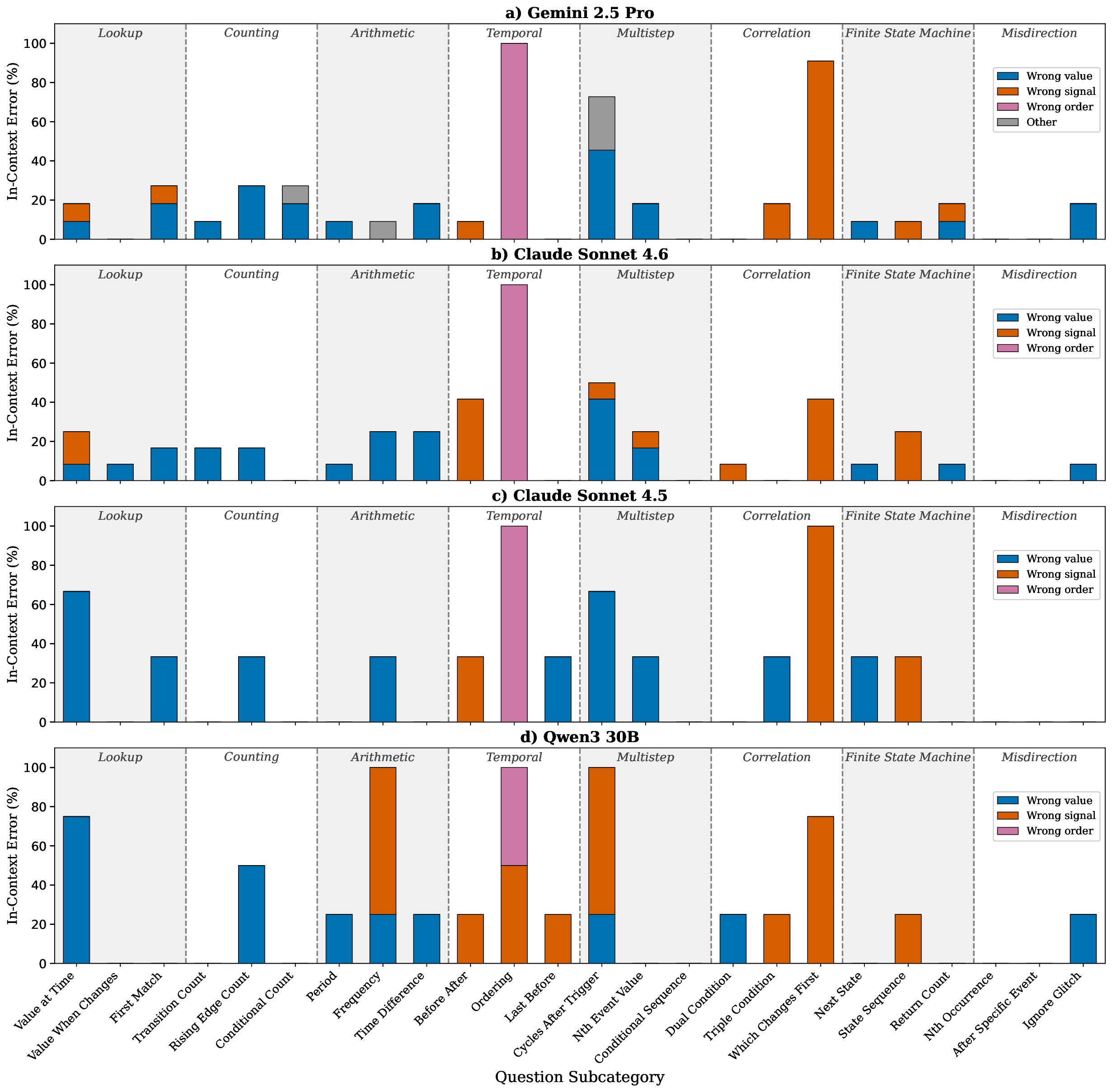}
    \caption{In-context error rate (\%) by question subcategory for each model individually: (a)~Gemini 2.5 Pro, (b)~Claude Sonnet 4.6, (c)~Claude Sonnet 4.5, (d)~Qwen3 30B. Bars are grouped by category and decomposed into error types: wrong value (blue), wrong signal (orange), wrong order (pink), and other (gray, panel~a only).}
    \label{fig:detailed_subtype_vs_transitions}
\end{figure*}

Figure~\ref{fig:detailed_subtype_vs_transitions} shows the per-model in-context error by subcategory, complementing the aggregated view in Figure~\ref{fig:error_analysis}.

\subsection{Question Generation Prompts}
\label{appendix:q_prompts}

Table~\ref{tab:generator-prompts} shows the question templates for all 24 subcategories. Some subcategories have multiple templates. These template variations are indicated by parentheses in the table.

\begin{table*}[t]
\caption{List of Prompts Created by Deterministic Generators}
\label{tab:generator-prompts}
\centering
\scriptsize
\setlength{\tabcolsep}{3pt}
\renewcommand{\arraystretch}{1.1}
\begin{tabular}{@{}p{2.4cm}p{6.0cm}p{2.4cm}p{6.0cm}@{}}
\toprule
\textbf{Sub-Category} & \textbf{Prompt} & \textbf{Sub-Category} & \textbf{Prompt} \\
\midrule
\multicolumn{4}{l}{\cellcolor{gray!15}\textbf{Lookup}} \\
\textbf{Value at Time (Indirect)} & "What is the value of \{signal.name\} at the \{ordinal\} rising edge of \{ref\_sig.name\}?" & \textbf{Value at Time (Direct)} & "What is the value of \{signal.name\} at t=\{time\}\{time\_unit\}?" \\
\textbf{First Match (Single)} & "What is the value of \{query\_sig.name\} the first time \{match\_sig.name\} equals \{target\_hex\}?" & \textbf{First Match (Two Cond.)} & "What is the value of \{query\_sig.name\} the first time \{cond1.name\}=\{value1\} AND \{cond2.name\}=\{value2\} simultaneously?" \\
\textbf{Value When Changes} & "What is the value of \{sig\_query.name\} when \{sig\_trigger.name\} transitions from \{from\_val\} to \{to\_val\} for the \{ordinal\} time?" & & \\
\midrule
\multicolumn{4}{l}{\cellcolor{gray!15}\textbf{Counting}} \\
\textbf{Transition Count} & "How many times does \{signal.name\} change value between t=\{t\_start\}\{time\_unit\} and t=\{t\_end\}\{time\_unit\}?" & \textbf{Rising Edge (Gated)} & "How many rising edges does \{signal.name\} have while \{gate\_signal.name\} is low, between t=\{start\}\{time\_unit\} and t=\{end\}\{time\_unit\} (inclusive)?" \\
\textbf{Rising Edge (Ungated)} & "How many rising edges does \{signal.name\} have between t=\{start\}\{time\_unit\} and t=\{end\}\{time\_unit\} (inclusive)?" & \textbf{Falling Edge Count} & "How many falling edges does \{signal.name\} have between t=\{start\}\{time\_unit\} and t=\{end\}\{time\_unit\} (inclusive)?" \\
\textbf{Conditional Count} & "How many rising edges does \{sig\_count.name\} have while \{sig\_cond.name\} is high, between t=\{win\_start\}\{time\_unit\} and t=\{win\_end\}\{time\_unit\}?" & & \\
\midrule
\multicolumn{4}{l}{\cellcolor{gray!15}\textbf{Arithmetic}} \\
\textbf{Period (Reference)} & "What is the period of \{signal.name\} between the \{lo\_ord\} and \{hi\_ord\} rising edges of \{ref.name\}, in \{time\_unit\}?" & \textbf{Period (Time Window)} & "What is the period of \{signal.name\} between t=\{win\_start\}\{time\_unit\} and t=\{win\_end\}\{time\_unit\}, in \{time\_unit\}?" \\
\textbf{Frequency (Reference)} & "What is the frequency of \{signal.name\} between the \{lo\_ord\} and \{hi\_ord\} rising edges of \{ref.name\}, in \{freq\_unit\}?" & \textbf{Frequency (Time Window)} & "What is the frequency of \{signal.name\} between t=\{t\_start\}\{time\_unit\} and t=\{t\_end\}\{time\_unit\}, in \{freq\_unit\}?" \\
\textbf{Time Difference} & "What is the time difference from the \{ord1\} \{edge\_type1\} edge of \{sig1.name\} to the \{ord2\} \{edge\_type2\} edge of \{sig2.name\}, in \{time\_unit\}?" & & \\
\midrule
\multicolumn{4}{l}{\cellcolor{gray!15}\textbf{Temporal}} \\
\textbf{Before/After} & "Does \{sig1.name\} change before, after, or at the same time as \{sig2.name\} \{reference\_description\}?" & \textbf{Ordering} & "List these signals in the order they first change after t=\{anchor\}\{time\_unit\}: \{shuffled\_names\}. Provide a comma-separated list." \\
\textbf{Last Before} & "What is the LAST value of \{data\_sig.name\} \{ref\_desc\}? Give the value immediately preceding the rising edge." & & \\
\midrule
\multicolumn{4}{l}{\cellcolor{gray!15}\textbf{Multistep}} \\
\textbf{Cycles After Trigger} & "What is the value of \{data.name\} exactly \{n\_cycles\} \{clock.name\} cycles after \{trigger\_phrase\}?" & \textbf{nth Event Value} & "What is the value of \{query\_sig.name\} immediately after the \{ordinal\} \{edge\_desc\} of \{event\_sig.name\}?" \\
\textbf{Cond. Sequence (3 Sig.)} & "What is the value of \{sig\_c.name\} when \{sig\_a.name\} goes high for the first time after the \{ordinal\} time \{sig\_b.name\} goes low?" & \textbf{Cond. Sequence (4 Sig.)} & "What is the value of \{sig\_c.name\} when \{sig\_a.name\} goes high for the first time after the \{ordinal\} time \{sig\_b.name\} goes low, while \{sig\_d.name\}=1?" \\
\midrule
\multicolumn{4}{l}{\cellcolor{gray!15}\textbf{Correlation}} \\
\textbf{Dual Condition} & "What is the value of \{query\_sig.name\} at the \{match\_ordinal\} time when \{cond1.name\}=\{v1\} AND \{cond2.name\}=\{v2\}?" & \textbf{Triple Condition} & "Between t=\{t\_win\_lo\}\{tu\} and t=\{t\_win\_hi\}\{tu\}, what is the value of \{query\_sig.name\} at the \{match\_ordinal\} time when \{cond1\}=\{val1\} AND \{cond2\}=\{val2\} AND \{cond3\}=\{val3\}?" \\
\textbf{Which Changes Last} & "After \{trigger.name\} goes high at t=\{trigger\_time\}\{time\_unit\}, which signal changes LAST: \{options\}?" & & \\
\midrule
\multicolumn{4}{l}{\cellcolor{gray!15}\textbf{Finite State Machine}} \\
\textbf{Next State} & "What state does \{signal.name\} transition to the \{ordinal\} time it leaves \{from\_display\}\{time\_filter\}?" & \textbf{State Sequence} & "List the sequence of states that \{signal.name\} visits between t=\{start\_time\}\{time\_unit\} and t=\{end\_time\}\{time\_unit\}. Provide as a comma-separated list." \\
\textbf{Return Count (Specific)} & "How many times does \{signal.name\} transition directly from \{from\_display\} to \{to\_display\}\{time\_window\}?" & \textbf{Return Count (Any)} & "How many times does \{signal.name\} transition INTO \{state\_display\}\{time\_window\}? (Count each entry, not time spent)" \\
\midrule
\multicolumn{4}{l}{\cellcolor{gray!15}\textbf{Misdirection}} \\
\textbf{nth Occurrence} & "\{trigger\_sig.name\} has multiple edges: \{times\_str\}, etc. What is the value of \{query\_sig.name\} immediately after the \{ordinal\} RISING edge of \{trigger\_sig.name\}?" & \textbf{After Specific Event} & "What is the value of \{query\_sig.name\} when \{trigger\_sig.name\} goes high for the first time AFTER the \{ref\_ordinal\} time \{ref\_sig.name\} goes low?" \\
\textbf{Ignore Glitch} & "Signal \{sig.name\} has \{pulse\_desc\}. Ignoring these transients, what is the stable value of \{sig.name\} during the interval t=\{interval\_start\} to t=\{interval\_end\}\{time\_unit\}?" & & \\
\bottomrule
\end{tabular}
\end{table*}

\end{document}